\documentclass[conference,10pt]{IEEEtran}

\usepackage[dvips]{graphicx}
\usepackage[english]{babel}
\usepackage[latin1]{inputenc}

\usepackage{amsmath}
\usepackage{amssymb}
\usepackage{amsbsy}
\usepackage{url}
\usepackage[english,ruled,vlined]{algorithm2e}
\usepackage{pstricks}

\usepackage{cite} % Loading the cite package will result in citation numbers being automatically sorted and properly "ranged". i.e., [1], [9], [2], [7], [5], [6] (without using cite.sty) will become: [1], [2], [5]--[7], [9] (using cite.sty)
\newcommand{\betP}{\mathrm{betP}}
\newcommand{\ind}{\mbox{1\hspace{-.25em}l}}

\newcommand{\conj}{\mathrm{Conj}}
\newcommand{\dis}{\mathrm{Dis}}
\newcommand{\mix}{\mathrm{Mix}}
\newcommand{\emix}{\mathrm{EMix}}
\newcommand{\size}{\mathrm{size}}
\newcommand{\lprod}{\mathrm{lprod}}
\newcommand{\lsum}{\mathrm{lsum}}
\newcommand{\DP}{\mathrm{DP}}
\newcommand{\Y}{\mathrm{Y}}
\newcommand{\DS}{\mathrm{DS}}
\newcommand{\Flo}{\mathrm{Flo}}
\newcommand{\PCR}{\mathrm{PCR}}
\newcommand{\DPCR}{\mathrm{DPCR}}
\newcommand{\MDPCR}{\mathrm{MDPCR}}
\newcommand{\PCRmo}{\mathrm{PCR6}}

\newcommand{\Pmof}{\mathrm{PCR6f}}
\newcommand{\Pmog}{\mathrm{PCR6g}}
\newcommand{\argmax}{\operatornamewithlimits{argmax}}

\begin{document}

\title{Toward a combination rule to deal with partial conflict and specificity in belief functions theory}
\author{\authorblockN{Arnaud Martin}
\authorblockA{$\mbox{E}^3\mbox{I}^2$ EA3876 \\
ENSIETA \\
2 rue Fran{\c c}ois Verny,\\
 29806 Brest Cedex 09, France \\
Email: Arnaud.Martin@ensieta.fr}
\and
\authorblockN{Christophe Osswald}
\authorblockA{$\mbox{E}^3\mbox{I}^2$ EA3876 \\
ENSIETA \\
2 rue Fran{\c c}ois Verny,\\
 29806 Brest Cedex 09, France \\
Email: Christophe.Osswald@ensieta.fr}}

\maketitle

\selectlanguage{english}

\begin{abstract}
We present and discuss a mixed conjunctive and disjunctive rule, a generalization of conflict repartition rules, and a combination of these two rules. In the belief functions theory one of the major problem is the conflict repartition enlightened by the famous Zadeh's example. To date, many combination rules have been proposed in order to solve a solution to this problem. Moreover, it can be important to consider the specificity of the responses of the experts. Since few year some unification rules are proposed. We have shown in our previous works the interest of the proportional conflict redistribution rule. We propose here a mixed combination rule following the proportional conflict redistribution rule modified by a discounting procedure. This rule generalizes many combination rules.
\end{abstract}


% for peerreview papers, inserts a page break and creates the second title.
% Will be ignored for other modes.
\IEEEpeerreviewmaketitle
%==================================================================
\section{Introduction}
\label{sec:Introduction}
Many fusion theories have been studied for the combination of the experts opinions such as voting rules \cite{Xu92,Lam97}, possibility theory \cite{Zadeh78,Dubois88a}, and belief functions theory \cite{Dempster67,Shafer76}. We can divide all these fusion approaches into four steps: \textit{modelization}, \textit{parameters estimation} depending on the model (not always necessary), \textit{combination}, and \textit{decision}. The most difficult step is presumably the first one. However, it is only at the combination step that we can add information such as the conflict between expert or the specificity of the expert's response. 

The voting rules are not adapted to the modelization of conflict between experts. If both possibility and probability-based theories can model imprecise and uncertain data at the same time, in a lot of applications experts can express their certainty on their perception of the reality. As a result, probabilities-based theory such as the belief functions theory is more adapted. 

The belief functions theory, also called evidence theory or Dempster-Shafer theory \cite{Dempster67, Shafer76} is based on the use of functions defined on the power set $2^\Theta$ (the set of all the subsets of $\Theta$), where $\Theta$ is the set of elements.  Theses \textit{belief functions} or \textit{basic belief assignments}, $m$ are defined by the mapping of the power set $2^\Theta$ onto $[0,1]$ with:

\begin{equation}
\label{close}
m(\emptyset)=0,
\end{equation}
and 
\begin{equation}
\label{normDST}
\sum_{X\in 2^\Theta} m(X)=1.
\end{equation}

The equation (\ref{close}) is the hypothesis of a closed world \cite{Shafer76}. We can define the belief function only with:
\begin{equation}
\label{open}
m(\emptyset)\geq 0,
\end{equation}
and the world is open \cite{Smets90}. In order to change an open world to a closed world, we can add one element in the discriminant space.

These simple conditions in equation (\ref{close}) and (\ref{normDST}), give a large panel of definitions of the belief functions, which is one the difficulties of the theory. From these basic belief assignments, other belief functions can be defined such as the credibility and the plausibility. To keep a maximum of information, it is preferable to combine information given by the basic belief assignments into a new basic belief assignment and take the decision on the obtained belief functions. If the credibility function provides a pessimistic decision, the plausibility function is often too optimistic. The pignistic probability \cite{Smets90} is generally considered as a compromise. It is given for all $X \in 2^\Theta$, with $X \neq \emptyset$ by:
\begin{eqnarray}
\label{pignistic}
\betP(X)=\sum_{Y \in 2^\Theta, Y \neq \emptyset} \frac{|X \cap Y|}{|Y|} \frac{m(Y)}{1-m(\emptyset)}.
\end{eqnarray}

The normalized conjunctive combination rule is the first rule proposed in the belief theory by \cite{Dempster67}. In the belief functions theory one of the major problem is the conflict repartition enlightened by the famous Zadeh's example. To date, many combination rules have been proposed, building a solution to this problem \cite{Yager87, Dubois88, Smets90b, Inagaki91, Smets93, Josang03, Smarandache05, Florea06, Martin06b, Denoeux06}. Last years some unification rules have been proposed \cite{Smets97, Lefevre02a, Appriou05}. 

The remainder of the paper is organized as follows. Section \ref{combination_rules} highlights the importance of the conflict in the classical combination rules. An historical point of view of the combination rules and the proportional conflict redistribution rules are recalled. A general formulation for combination rules is presented in Section \ref{general_formulation}. First we propose a mixed rule between the conjunctive and disjunctive rules in subsection \ref{Mix_conj_dis}, and a proportional conflict repartition rules with a discounting procedure in subsection \ref{general_PCR}. From these two new rules we propose a more general rule in subsection \ref{General_rule}. Section \ref{illustration} presents a discussion for a more general rule, and finally the section \ref{conclusion} outlines the conclusions of the paper. An algorithm implementation is proposed in section \ref{s_algo}.

\section{The classical combination rules and the conflict repartition}
\label{combination_rules}

\subsection{An historical point of view}

The first combination rule proposed by Dempster and Shafer is the normalized conjunctive combination rule given for two basic belief assignments $m_1$ and $m_2$ and for all $X \in 2^\Theta$, $X\neq \emptyset$ by:
\begin{eqnarray}
m_\DS(X)=\displaystyle \frac{1}{1-k}\sum_{A\cap B =X} m_1(A)m_2(B),
\end{eqnarray}
where $k= \displaystyle \sum_{A\cap B =\emptyset} m_1(A)m_2(B)$ is the global conflict of the combination. The problem enlightened by the famous Zadeh's example is the repartition of the global conflict. Indeed, consider $\Theta=\{A,B,C\}$ and two experts opinions given by $m_1(A)=0.9$, $m_1(C)=0.1$, and $m_2(B)=0.9$, $m_1(C)=0.1$, the mass given by the combination is $m(C)=1$. 

So as to resolve this problem Smets \cite{Smets90b} proposes to consider an open world, therefore the conjunctive rule is non-normalized and we have for two basic belief assignments $m_1$ and $m_2$ and for all $X \in 2^\Theta$ by:
\begin{eqnarray}
\label{conjunctive}
m_\conj(X)=\displaystyle \sum_{A\cap B =X} m_1(A)m_2(B).
\end{eqnarray}
$m_\conj(\emptyset)$ can be interpreted as a non-expected solution. In the Transferable Belief Model of Smets, the repartition of the global conflict is done in the decision step by the pignisitic probability (\ref{pignistic}).

Yager \cite{Yager87} proposes to transfer the global conflict on the ignorance $\Theta$:
\begin{eqnarray}
\label{Yager}
\begin{array} {rcl}
m_\Y(X)&=& m_\conj(X), \forall X \in 2^\Theta \smallsetminus \{\emptyset, \Theta\} \\
m_\Y(\Theta)&=&m_\conj(\Theta)+m_\conj(\emptyset)\\
m_\Y(\emptyset)&=&0.
\end{array}
\end{eqnarray}

These three based-conjunctive rules reduce the imprecision and uncertainty but can be used only if all the experts are reliable. In the other case a disjunctive combination rule can be used \cite{Smets93} given for two basic belief assignments $m_1$ and $m_2$ and for all $X \in 2^\Theta$ by:
\begin{eqnarray}
m_\dis(X)=\displaystyle \sum_{A\cup B =X} m_1(A)m_2(B).
\end{eqnarray}
Of course with this rule we have a loss of specificity.

When we can quantify the reliability of each expert, we can weaken the basic belief assignment before the combination by the discounting procedure:
\begin{eqnarray}
\left\{
\begin{array}{l}
	m'_j(X)=\alpha_j m_j(X), \, \forall X \in 2^\Theta \smallsetminus \{\Theta\} \\
	m'_j(\Theta)=1-\alpha_j (1-m_j(\Theta)).
\end{array}
\right.
\end{eqnarray}
$\alpha_j \in [0,1]$ is the discounting factor of the expert $j$ that is in this case the reliability of the expert $j$, eventually as a function of $X \in 2^\Theta$.

Dubois and Prade \cite{Dubois88} propose a mixed rule with a repartition of the partial conflict on the partial ignorance. Consequently, the conflict is considered more precisely than previously. This rule is given for two basic belief assignments $m_1$ and $m_2$ and for all $X \in 2^\Theta$, $X\neq \emptyset$ by:

\begin{eqnarray}
\label{DP}
\begin{array}{rcl}
m_\DP(X)&=&\displaystyle \sum_{A \cap B = X} m_1(A) m_2(B)\\
&&+\displaystyle \sum_{
\begin{array}{c}
\scriptstyle A \cup B=X\\
\scriptstyle A \cap B = \emptyset \\
\end{array}} m_1(A)m_2(B).
\end{array}
\end{eqnarray}

The repartition of the conflict is important because of the non-idempotency of the rules (except the rule of \cite{Denoeux06} that can be applied when the dependency between experts is high) and due to the responses of the experts that can be conflicting. Hence, we have define the auto-conflict \cite{Osswald06} in order to quantify the intrinsic conflict of a mass and the distribution of the conflict according to the number of experts.

\subsection{The proportional conflict redistribution rules}
Dezert and Smarandache proposed a list of proportional conflict redistribution ($\PCR$) methods \cite{Smarandache05,Smarandache06} to redistribute the partial
conflict on the elements implied in the partial conflict. 

The most efficient is the $\PCR$ rule given for two basic belief assignments $m_1$ and $m_2$ and for all $X \in 2^\Theta$, $X\neq \emptyset$ by:
\begin{eqnarray}
\label{DSmTcombination}
\begin{array}{l}
m_\PCR(X)=m_\conj(X)~+\\
\displaystyle
\sum_{\begin{array}{l}
\scriptstyle Y\in 2^\Theta, \\
\scriptstyle X\cap Y = \emptyset 
\end{array}} \!\!\!\!\!\left(\frac{m_1(X)^2 m_2(Y)}{m_1(X) \!+\!
  m_2(Y)}+\frac{m_2(X)^2 m_1(Y)}{m_2(X) \!+\! m_1(Y)}\!\right)\!\!,
\end{array}
\end{eqnarray}
where $m_\conj(.)$ is the conjunctive rule given by the equation (\ref{conjunctive}). We have studied and formulated this rule for more than two experts in \cite{Martin06b, Osswald06} $X \in 2^\Theta$, $X\neq \emptyset$:

\begin{eqnarray}
\label{PCR6combination}
\begin{array}{c}
  \displaystyle m_\PCRmo(X)  =  \displaystyle m_\conj(X) + \sum_{i=1}^M
  m_i(X)^2 \\
  \!\!\!\!\!\!\!\! \displaystyle \sum_{\begin{array}{c}
      \scriptstyle {\displaystyle \mathop{\cap}_{k=1}^{M\!-\!1}} Y_{\sigma_i(k)} \cap X = \emptyset \\
      \scriptstyle (Y_{\sigma_i(1)},...,Y_{\sigma_i(M\!-\!1)})\in (2^\Theta)^{M\!-\!1}
  \end{array}}
  \!\!\!\!\!\!\!\!\!\!\!\!
  \left(\!\!\frac{\displaystyle \prod_{j=1}^{M\!-\!1} m_{\sigma_i(j)}(Y_{\sigma_i(j)})}
       {\displaystyle m_i(X) \!+\! \sum_{j=1}^{M\!-\!1} m_{\sigma_i(j)}(Y_{\sigma_i(j)})}\!\!\right)\!\!,
\end{array}
\end{eqnarray}
where $Y_j \in 2^\Theta$ is the response of the expert $j$, $m_j(Y_j)$ the associated belief function and $\sigma_i$ counts from 1 to $M$ avoiding $i$:
\begin{eqnarray}
\label{sigma}
\left\{
\begin{array}{ll}
\sigma_i(j)=j &\mbox{if~} j<i,\\
\sigma_i(j)=j+1 &\mbox{if~} j\geq i,\\
\end{array}
\right.
\end{eqnarray}

The idea is here to redistribute the masses of the focal elements giving a partial conflict proportionally to the initial masses on these elements.

\section{A general formulation for combination rules}
\label{general_formulation}

In \cite{Smets97, Appriou05} we can find two propositions of a general formulation of the combination rules. In the first one, Smets considers the combination rules from a matrix notation and find the shape of this matrix according to some assumptions on the rule, such as linearity, commutativity, associativity, etc. In the second one, a generic operator is defined from the plausibility functions. 

A general formulation of the global conflict repartition have been proposed in \cite{Inagaki91, Lefevre02a} for all $X\in 2^\Theta$ by:
\begin{eqnarray}
\label{rep}
m_c(X)=m_\conj(X)+w(X)m_\conj(\emptyset),
\end{eqnarray}
where $\displaystyle \sum_{X\in 2^\Theta} w(X)=1$. The problem is the choice of the weights $w(X)$.

We have proposed also a parametrized $\PCR$ to decrease or increase the influence of many small values toward one large one. The first way is given by
$\Pmof$, applying a function on each belief value implied in the partial conflict. Any non-decreasing positive
function $f$ defined on $]0,1]$ can be used.

\begin{eqnarray*}
  \label{PCR6mof}
  \begin{array}{c}
    \displaystyle m_{\Pmof}(X)  =  \displaystyle m_\conj(X) + \sum_{i=1}^M
    m_i(X) f(m_i(X)) \\
   \displaystyle \sum_{\begin{array}{c}
	\scriptstyle {\displaystyle \mathop{\cap}_{k=1}^{M\!-\!1}} Y_{\sigma_i(k)} \cap X = \emptyset \\
	\scriptstyle (Y_{\sigma_i(1)},...,Y_{\sigma_i(M\!-\!1)})\in (2^\Theta)^{M\!-\!1}
    \end{array}}
    \!\!\!\!\!\!\!\!\!\!\!\!
    \left(\!\!\frac{\displaystyle \prod_{j=1}^{M\!-\!1} m_{\sigma_i(j)}(Y_{\sigma_i(j)})}
	 {\displaystyle f(m_i(X)) \!+\! \sum_{j=1}^{M\!-\!1} m_{\sigma_i(j)}f(Y_{\sigma_i(j)})}\!\!\right)
  \end{array}
\end{eqnarray*}

The second way, given by $\Pmog$ is to apply a similar function $g$ on the sum of belief functions given to a focal element. 

\begin{eqnarray*}
  \label{PCR6mog}
  \begin{array}{c}
  \displaystyle m_{\Pmog}\!(X) \!=\! m_\conj(X) \!+\!\! \sum_{i=1}^M
  m_i(X) \!\!\!\!\!\!\!\!\!\!\!\!\!\!\!\!\!\!\!\!\! \sum_{\begin{array}{c}
      \scriptstyle {\displaystyle \mathop{\cap}_{\scriptscriptstyle
	  k=1}^{\scriptscriptstyle M\!-\!1}} Y_{\sigma_i(k)} \cap X
      = \emptyset \\
      \scriptstyle (Y_{\sigma_i(1)},...,Y_{\sigma_i(M\!-\!1)})\in (2^\Theta)^{M\!-\!1}
  \end{array}} \\
  
    \frac{\displaystyle
      \!\!\!\!\!\!\!\!\Bigg(\!\prod_{j=1}^{M\!-\!1}\!\!m_{\sigma_i(j)}(Y_{\sigma_i(j)})\!\!\!\Bigg)
      \!\!\Bigg(\!\!\!\!\!\!\!\prod_{~~~Y_{\sigma_i(j)}=X}\!\!\!\!\!\!\!\!\!\!\!\ind_{j>i}\!\!\Bigg)
      g\Bigg(\!\!m_i(X)\!\!+\!\!\!\!\!\!\!\!\sum_{Y_{\sigma_i(j)}=X}\displaystyle \!\!\!\!\!\!\!\!m_{\sigma_i(j)}(Y_{\sigma_i(j)})\!\!\!\Bigg)}
	 {\displaystyle \!\!\!\!\!\!\!\!\!\!\!\!\sum_{~~~~~\renewcommand{\arraystretch}{2}\begin{array}{c}
	       \scriptstyle Z\in\{X, Y_{\sigma_i(1)}, \ldots, Y_{\sigma_i(M\!-\!1)}\}
	   \end{array}}
	   \!\!\!\!\!\!\!\!\!\!\!\!\!\!\! g\left(\!\sum_{Y_{\sigma_i(j)}=Z}\!\!\!\!\!\displaystyle m_{\sigma_i(j)}(Y_{\sigma_i(j)})+m_i(X)\ind_{\scriptscriptstyle X=Z}\!\!\!\right)}
  \end{array}
\end{eqnarray*}
where $\ind$ is the characteristic function ($\ind_{\scriptscriptstyle X=Z}$ is 1 if $X=Z$ and 0 elsewhere). Nevertheless, here also the problem is the choice of the functions $f$ and $g$. 

\subsection{How to choose conjunctive and disjunctive rules?}
\label{Mix_conj_dis}

We have seen that conjunctive rule reduces the imprecision and uncertainty but can be used only if one of the experts is reliable, whereas the disjunctive rule can be used when the experts are not reliable, but allows a loss of specificity. 

Hence, Florea \cite{Florea06} proposes a weighted sum of these two rules according to the global conflict $k=m_\conj(\emptyset)$ given for $X \in 2^\Theta$, $X\neq \emptyset$ by:
\begin{eqnarray}
\label{Florea}
m_\Flo(X)=\displaystyle \beta_1(k) m_\dis(X)+ \beta_2(k) m_\conj(X),
\end{eqnarray}
where $\beta_1$ and $\beta_2$ can admit $\displaystyle k=\frac{1}{2}$ as symmetric weight: 
\begin{eqnarray}
\begin{array}{l}
\beta_1(k)=\displaystyle \frac{k}{1-k+k^2},\\
\beta_2(k) = \displaystyle \frac{1-k}{1-k+k^2}.\\
\end{array}
\end{eqnarray}
Consequently, if the global conflict is high ($k$ near 1) the behavior of this rule will give more importance to the disjunctive rule. Thus, this rule considers the global conflict coming from the non-reliability of the experts. 

In order to take into account the weights more precisely in each partial combination, we propose the following new rule. For two basic belief assignments $m_1$ and $m_2$ and for all $X \in 2^\Theta$ we have:
\begin{eqnarray}
\begin{array}{rcl}
m_\mix(X)&=&\displaystyle \sum_{A\cup B =X} \delta_1 m_1(A)m_2(B)\\
&+& \displaystyle \sum_{A\cap B =X} \delta_2 m_1(A)m_2(B).
\end{array}
\end{eqnarray}
Of course, if $ \delta_1=\beta_1(k)$ and $\delta_2=\beta_2(k)$ we obtain the Florea's rule. In the same manner, if $\delta_1=1- \delta_2=0$ we obtain the conjunctive rule and if $\delta_1=1- \delta_2=1$ the disjunctive rule. If $\delta_1(A,B)=1- \delta_2(A,B)=\ind_{A\cap B =\emptyset}$ we retrieve the Dubois and Prade's rule and the partial conflict can be considered, whereas the rule (\ref{Florea}).

The choice of $\delta_1=1- \delta_2$ can be done by a dissimilarity such as: 
\begin{eqnarray}
\label{eq_delta2}
\delta(A,B)=\displaystyle 1-\frac{|A\cap B|}{\min (|A|, |B|)},
\end{eqnarray}
where $|A|$ is the cardinality of $A$. Note that is not a distance nor a proper dissimilarity (\emph{e.g.} $\delta(A,B)=0$ does not imply $A=B$). We can also take for $\delta_2$, the Jaccard distance given by:
\begin{eqnarray}
\label{eq_d2}
d(A,B)=\displaystyle \frac{|A\cap B|}{|A\cup B|},
\end{eqnarray}
used by \cite{Jousselme01} on the belief functions. Thus, if we have a partial conflict between $A$ and $B$, $|A\cap B|=0$ and the rule transfers the mass on $A\cup B$. In the case $A\subset B$ (or the contrary), $A\cap B=A$ and $A \cup B=B$, so with $\delta$ the rule transfers the mass on $A$ and with $d$ on $A$ and $B$ according to the rate $|A|/|B|$ of the cardinalities. In the case  $A\cap B \neq A, B$ and $\emptyset$, the rule transfers the mass on $A \cap B$ and $A \cup B$ according to $\delta$ and $d$.

Consider the following example for two experts on \linebreak $\Theta=\{A,B,C\}$:
\begin{center}
  \begin{tabular}{|l|c|c|c|c|c|c|}
    \hline
    & $\emptyset$ & $A$ & $B$ & $A\cup B$ & $A\cup C$ & $\Theta$ \\
    \hline
    Expert 1  & 0& 0.3 & 0 & 0.4 & 0 & 0.3 \\
    \hline
    Expert 2 & 0 & 0 & 0.2 & 0 & 0.5 & 0.3 \\
    \hline
    $m_\conj$ & 0.06 & 0.44 & 0.14 & 0.12 & 0.15 & 0.09 \\
    \hline
  \end{tabular}
\end{center}
Accordingly, we obtain for $\delta_{1}=\delta$:
\begin{center}
  \begin{tabular}{|c|c|c|c|}
    \hline
    $\delta$ &  $A$ &  $A\cup B$ & $\Theta$ \\
    \hline
    $B$  & 1& 0 & 0  \\
    \hline
    $A\cup C$ & 0 & 1/2 & 0 \\
    \hline
    $\Theta$ & 0 & 0 & 0 \\
    \hline
  \end{tabular}
\end{center}
where the columns are the focal elements of the basic belief assignment given by the expert 1 and the lines are the focal elements of the basic belief assignment given by expert 2. The mass 0.06 on $\emptyset$ is transfered on $A\cup B$ and the mass 0.2 on $A$ given by the responses $A\cup B$ and $A\cup C$ is transfered on $A$ with a value of 0.1 and on $\Theta$ with the same value. For $\delta_1=1-d$ we have:
\begin{center}
  \begin{tabular}{|c|c|c|c|}
    \hline
    $1-d$ &  $A$ &  $A\cup B$ & $\Theta$ \\
    \hline
    $B$  & 1& 1/2 & 2/3  \\
    \hline
    $A\cup C$ & 1/2 & 2/3 & 1/3 \\
    \hline
    $\Theta$ & 2/3 & 1/3 & 0 \\
    \hline
  \end{tabular}
\end{center}
Note that $\delta$ can be used when the experts are considered reliable. In  this case we consider the more precise response. With $d$, we have the conjunctive rule only when the experts give the same response, else we consider the doubtful responses and we transfer the masses in proportion of the imprecision of the responses (given by the cardinality of the responses) on the part in agreement and on the partial ignorance. 

For more than two experts, if the intersection of the responses of the experts is not empty, we can still transfer on the intersection and the union, and the equation (\ref{eq_delta2}) becomes:
\begin{eqnarray}
\label{delta}
\delta(Y_1,...,Y_M)=\displaystyle 1-\frac{|Y_1 \cap ... \cap Y_M|}{\displaystyle \min_{1\leq i \leq M}|Y_i|}.
\end{eqnarray} 
From the equation (\ref{eq_d2}), we can define $\delta$ by:
\begin{eqnarray}
\label{d}
\delta(Y_1,...,Y_M)=\displaystyle 1-\frac{|Y_1 \cap ... \cap Y_M|}{|Y_1 \cup ... \cup Y_M|}.
\end{eqnarray} 
Finally, the rule is given by:
\begin{eqnarray}
\label{mixed}
\begin{array}{l}
m_\mix(X)=\displaystyle \sum_{Y_1 \cup ... \cup Y_M=X} \delta(Y_1,...,Y_M) \prod_{j=1}^M m_j(Y_j)\\
\quad \quad \quad + \displaystyle \sum_{Y_1 \cap ... \cap Y_M=X} (1-\delta(Y_1,...,Y_M)) \prod_{j=1}^M m_j(Y_j).
\end{array}
\end{eqnarray}

This formulation can be interesting according to the coherence of the responses of the experts. However, this formulation does not allow the repartition of the partial conflict in an other way than the Dubois and Prade's rule. In the later (\ref{MDPCR2}) and (\ref{eq_mixed_to_dpcr}) equations, we will simply write $\delta$ instead of $\delta(Y_1,...,Y_M)$.

\subsection{A discounting proportional conflict repartition rule}
\label{general_PCR}

The $\PCRmo$ redistributes the masses of the conflicting focal elements proportionally to the initial masses on these elements. For instance, consider three experts expressing their opinion on $\Theta=\{A,B,C,D\}$:
\begin{center}
  \begin{tabular}{|l|c|c|c|c|}
    \hline
    & $A$ & $B$ & $A\cup C$ & $\Theta$ \\
    \hline
    Expert 1  & 0.7 & 0 & 0 & 0.3 \\
    \hline
    Expert 2 & 0 & 0.5 & 0 & 0.5 \\
    \hline
    Expert 3 & 0 & 0 & 0.6 & 0.4 \\
    \hline
  \end{tabular}
\end{center}

The global conflict is given here by 0.21+0.14+0.09=0.44, coming from $A$, $B$ and $A\cup C$ for 0.21, $A$, $B$ and $\Theta$ for 0.14 and $\Theta$, $B$ and $A\cup C$ for 0.09. With the generalized $\PCRmo$ rule (\ref{PCR6combination}) we obtain:
{\renewcommand{\arraystretch}{2}
$$
\begin{array}{l}
\displaystyle m_\PCRmo(A) = 0.14+0.21+0.21\, \frac{7}{18}+0.14 \, \frac{7}{16} \simeq 0.493,\\
\displaystyle m_\PCRmo(B) = 0.06+0.21\, \frac{5}{18} +0.14\, \frac{5}{16}+0.09\, \frac{5}{14} \simeq 0.194,\\
\displaystyle m_\PCRmo(A\cup C) = 0.09+0.21\, \frac{6}{18}+0.09\, \frac{6}{14} \simeq 0.199,\\
\displaystyle m_\PCRmo(\Theta) = 0.06+0.14\, \frac{4}{16}+0.09\, \frac{3}{14} \simeq 0.114. \\
\end{array}
$$
}

First of all, the repartition is only on the elements given the partial conflict. We can apply a discounting procedure in the combination rule in order to transfer a part of the partial conflict on the partial ignorance. This new discounting $\PCR$ (noted $\DPCR$) can be expressed for two basic belief assignments $m_1$ and $m_2$ and for all $X \in 2^\Theta$, $X\neq \emptyset$ by:
\begin{eqnarray}
\begin{array}{l}
m_\DPCR(X)=m_\conj(X)\\

+ 
\displaystyle
\sum_{
\begin{array}{l}
\scriptstyle Y\in 2^\Theta, \\
\scriptstyle X\cap Y = \emptyset 
\end{array}} \alpha \left(\frac{m_1(X)^2 m_2(Y)}{m_1(X) \!+\!
  m_2(Y)}+\frac{m_2(X)^2 m_1(Y)}{m_2(X) \!+\! m_1(Y)}\!\right)\\

+
\displaystyle
\sum_{
\begin{array}{c}
\scriptstyle Y_1\cup Y_2 =X\\
\scriptstyle Y_1\cap Y_2 = \emptyset 
\end{array}}(1-\alpha)  m_1(Y_1)m_2(Y_2),\\
\end{array}
\end{eqnarray}
with $\alpha\in [0,1]$, the discounting factor.

In a general case for $M$ experts, we could write this rule as:
\begin{eqnarray}
\label{DPCR}
\begin{array}{l}
  \displaystyle m_\DPCR(X)  =  \displaystyle m_\conj(X) + \sum_{i=1}^M
  m_i(X)^2 \\
  \!\!\!\!\!\!\!\! \displaystyle \sum_{\begin{array}{c}
      \scriptstyle {\displaystyle \mathop{\cap}_{k=1}^{M\!-\!1}} Y_{\sigma_i(k)} \cap X = \emptyset \\
      \scriptstyle (Y_{\sigma_i(1)},...,Y_{\sigma_i(M\!-\!1)})\in (2^\Theta)^{M\!-\!1}
  \end{array}}
  \!\!\!\!\!\!\!\!\!\!\!\!
  \alpha \left(\!\!\frac{\displaystyle \prod_{j=1}^{M\!-\!1} m_{\sigma_i(j)}(Y_{\sigma_i(j)})}
       {\displaystyle m_i(X) \!+\! \sum_{j=1}^{M\!-\!1} m_{\sigma_i(j)}(Y_{\sigma_i(j)})}\!\!\right)\\

+
\displaystyle
\sum_{
\begin{array}{c}
\scriptstyle Y_1 \cup ... \cup Y_M = X\\
\scriptstyle Y_1 \cap ... \cap Y_M = \emptyset \\
\end{array}} (1-\alpha) \prod_{j=1}^M m_j(Y_j),
\end{array}
\end{eqnarray}
where $Y_j \in 2^\Theta$ is a response of the expert $j$, $m_j(Y_j)$ its assigned mass and $\sigma_i$ is given by (\ref{sigma}).

Hence, if we choose $\alpha=0.9$ in the previous example, we obtain:
{\renewcommand{\arraystretch}{2}
$$
\!\!\!\begin{array}{l}
\displaystyle m_\DPCR(A) = 0.14+0.21+0.21 \,  \frac{7}{18}\,  0.9+0.14\,  \frac{7}{16}\,  0.9\\
\quad \quad \quad \quad \quad \simeq 0.479,\\
\displaystyle m_\DPCR(B) = 0.06+0.21\,  \frac{5}{18}\,  0.9+0.14\,  \frac{5}{16}\,  0.9\\
\displaystyle \quad \quad \quad \quad \quad +0.09\,  \frac{5}{14}\,  0.9 \simeq 0.181,\\
\displaystyle m_\DPCR(A\cup C) = 0.09+0.21\,  \frac{6}{18}\,  0.9+0.09\,  \frac{6}{14}\,  0.9 \simeq 0.187,\\
\displaystyle m_\DPCR(A\cup B \cup C)= 0.21 \times  0.1 = 0.021,\\
\displaystyle m_\DPCR(\Theta) = 0.06+0.14\,  \frac{4}{16}\,  0.9+0.09\,  \frac{3}{14}\,  0.9+ 0.14 \times  0.1 \\
\displaystyle \quad \quad \quad  \quad + 0.09 \times  0.1 \simeq 0.132.\\
\end{array}
$$
}

However, in this example, the partial conflict due to the experts 1, 2 and 3 saying respectively $A$, $B$, and $A \cup C$, the conflict is 0.21; nonetheless only the experts 1 and 2 and the expert 2 and 3 are in conflict. The experts 1 and 3 are not in conflict. Now, consider another case where the experts 1, 2 and 3 say respectively $A$, $B$, and $C$ with the same conflict 0.21. In these both cases, the $\DPCR$ rule transfers the masses with the same weight $\alpha$. Although, we could prefer transfer more mass on $\Theta$ in the second than in the first case. 

Consequently, the transfer of mass can depend on the existence of conflict between each pair of experts. We define the conflict function giving the number of experts in conflict two by two for each response $Y_i \in 2^\Theta$ of the expert $i$ as the number of responses of the other experts in conflict with $i$. A function $f_i$ is defined by the mapping of $(2^\Theta)^M$ onto $\left[0,\displaystyle \frac{1}{M}\right]$ with:
\begin{eqnarray}
\label{f_function}
f_i(Y_1,...,Y_M)=\displaystyle \frac{\displaystyle  \sum_{j=1}^M\ind_{\{Y_j\cap Y_i= \emptyset\}}}{M(M-1)}.
\end{eqnarray}

Hence, we can choose $\alpha$ depending on the response of the experts such as:
\begin{eqnarray}
\label{alpha}
\alpha(Y_1,...,Y_M)=1-\sum_{i=1}^M f_i(Y_1,...,Y_M).
\end{eqnarray}
In this case $\alpha \in [0,1]$, likewise we do not transfer the mass on elements that can be written as the union of the responses of the experts.

Therefore, if we take again our previous example we obtain:
$$\alpha(A,B, A\cup C)=1-\frac{2}{3}=\frac{1}{3}, \alpha(A,B, \Theta)=1-\frac{1}{3}=\frac{2}{3},$$
$$\alpha(\Theta,B, A\cup C)=1-\frac{1}{3}=\frac{2}{3}.$$
Thus the provided mass by the $\DPCR$ is:
{\renewcommand{\arraystretch}{2}
$$
\!\!\! \begin{array}{l}
\displaystyle m_\DPCR(A) = 0.14+0.21+0.21 \, \frac{7}{18}\,  \frac{1}{3}+0.14\,  \frac{7}{16} \, \frac{2}{3} \simeq 0.418,\\
\displaystyle m_\DPCR(B) = 0.06+0.21 \, \frac{5}{18}\,  \frac{1}{3}+0.14 \, \frac{5}{16} \, \frac{2}{3}+0.09 \, \frac{5}{14} \, \frac{2}{3} \\
\displaystyle \quad \quad \quad \quad  \simeq 0.130,\\
\displaystyle m_\DPCR(A\cup C) = 0.09+0.21 \, \frac{6}{18} \, \frac{1}{3}+0.09 \, \frac{6}{14} \, \frac{2}{3}\simeq 0.139,\\
\displaystyle m_\DPCR(A\cup B \cup C)= 0.21 \, \frac{2}{3} = 0.140,\\
\displaystyle m_\DPCR(\Theta) = 0.06+0.14 \, \frac{4}{16} \, \frac{2}{3}+0.09 \, \frac{3}{14} \, \frac{2}{3}+ 0.14 \, \frac{1}{3} \\
\displaystyle \quad \quad \quad \quad  + 0.09 \, \frac{1}{3} \simeq 0.173.\\
\end{array}
$$
}

We want to take account of the degree of conflict (or non-conflict) within each pair of expert differently for each element. We can consider the non-conflict function given for each expert $i$ by the number of experts not in conflict with $i$. Hence, we can choose $\alpha_i(Y_1,...,Y_M)$ defined by the mapping of $(2^\Theta)^M$ onto $\left[\displaystyle 0, \frac{1}{M} \right]$ with:
\begin{eqnarray}
\label{g_function}
\begin{array}{rcl}
\alpha_i(Y_1,...,Y_M)&=&\displaystyle \frac{1}{M}- f_i(Y_1,...,Y_M)\\
&=& \displaystyle \frac{ \displaystyle  \sum_{j=1, j\neq i}^M\ind_{\{Y_j\cap Y_i \neq \emptyset\}}}{M (M-1)}.
\end{array}
\end{eqnarray}

The discounting PCR rule (equation (\ref{DPCR})) can be written for $M$ experts as:
\begin{eqnarray}
\label{DPCRi}
\begin{array}{l}
  \displaystyle m_\DPCR(X)  =  \displaystyle m_\conj(X) + \sum_{i=1}^M
  m_i(X)^2 \\
  \!\!\!\!\!\!\!\! \displaystyle \sum_{\begin{array}{c}
      \scriptstyle {\displaystyle \mathop{\cap}_{k=1}^{M\!-\!1}} Y_{\sigma_i(k)} \cap X = \emptyset \\
      \scriptstyle (Y_{\sigma_i(1)},...,Y_{\sigma_i(M\!-\!1)})\in (2^\Theta)^{M\!-\!1}
  \end{array}}
  \!\!\!\!\!\!\!\!\!\!\!\! 
  \alpha_i \lambda \left(\!\!\frac{\displaystyle \prod_{j=1}^{M\!-\!1} m_{\sigma_i(j)}(Y_{\sigma_i(j)})}
       {\displaystyle m_i(X) \!+\! \sum_{j=1}^{M\!-\!1} m_{\sigma_i(j)}(Y_{\sigma_i(j)})}\!\!\right)\\

+
\displaystyle
\sum_{
\begin{array}{c}
\scriptstyle Y_1 \cup ... \cup Y_M = X\\
\scriptstyle Y_1 \cap ... \cap Y_M = \emptyset \\
\end{array}} (1-\sum_{i=1}^M \alpha_i) \prod_{j=1}^M m_j(Y_j),
\end{array}
\end{eqnarray}
where $\alpha_i(X,Y_{\sigma_i(1)},...,Y_{\sigma_i(M-1)})$ is noted $\alpha_i$ for notations and $\lambda$ depending on $(X,Y_{\sigma_i(1)},...,Y_{\sigma_i(M-1)})$, is chosen to obtain the normalization given by the equation (\ref{normDST}). $\lambda$ is given when $\alpha_i\neq 0$, \linebreak $\forall i \in \{1,...,M\}$ by:

\begin{eqnarray}
\lambda=\frac{\displaystyle \sum_{i=1}^M \alpha_i}{<\boldsymbol{\alpha},\boldsymbol{\gamma}>},
\end{eqnarray}
where $<\boldsymbol{\alpha},\boldsymbol{\gamma}>$ is the scalar product of $\boldsymbol{\alpha}=(\alpha_i)_{i\in \{1,...,M\}}$ and $\boldsymbol{\gamma}=(\gamma_i)_{i\in \{1,...,M\}}$ with:
\begin{eqnarray}
\gamma_i = \frac{\displaystyle m_i(X)}
       {\displaystyle m_i(X) \!+\! \sum_{j=1}^{M\!-\!1} m_{\sigma_i(j)}(Y_{\sigma_i(j)})},
\end{eqnarray}
where $\gamma_i(X,Y_{\sigma_i(1)},...,Y_{\sigma_i(M-1)})$ is noted $\gamma_i$ for notations.

With this last version of the rule, for $\alpha_i$ given by the equation (\ref{g_function}), we obtain on our illustrative example  $\lambda=\frac{36}{13}$ when the experts 1, 2 and 3 say respectively $A$, $B$, and $A \cup C$ (the conflict is 0.21), $\lambda=\frac{16}{5}$ when the conflict is 0.14 and $\lambda=\frac{56}{17}$ when the conflict is 0.09. Thus, the masses are given by:
{\renewcommand{\arraystretch}{2}
$$
\begin{array}{l}
\displaystyle m_\DPCR(A) = 0.14+0.21+0.21 \, \frac{7}{18} \, \frac{1}{6} \, \frac{36}{13}+0.14 \, \frac{7}{16} \, \frac{1}{6} \, \frac{16}{5}\\
\quad \quad \quad \quad \quad \simeq 0.420,\\
\displaystyle m_\DPCR(B) = 0.06+ 0.14 \, \frac{5}{16} \, \frac{1}{6} \, \frac{16}{5} +0.09 \, \frac{5}{14} \, \frac{1}{6} \, \frac{56}{17} \simeq 0.101,\\
\displaystyle m_\DPCR(A\cup C) = 0.09+0.21 \, \frac{6}{18} \, \frac{1}{6} \, \frac{36}{13}+0.09 \, \frac{6}{14} \, \frac{1}{6} \, \frac{56}{17} \\
\quad \quad \quad \quad \quad \simeq 0.143,\\
\displaystyle m_\DPCR(A\cup B \cup C)= 0.21 \, \frac{2}{3} =0.14 \\
\displaystyle m_\DPCR(\Theta) = 0.06+0.14 \, \frac{4}{16} \, \frac{1}{3} \, \frac{16}{5}+0.09 \, \frac{3}{14} \, \frac{1}{3} \, \frac{56}{17} \\
\displaystyle \quad \quad \quad \quad + 0.14 \, \frac{1}{3}+ 0.09 \, \frac{1}{3} \simeq 0.196.\\
\end{array}
$$
}
This last version allows to consider a kind of degree of conflict (a degree of pair of non-conflict), but this degree is not so easy to introduce in the combination rule.

\subsection{A mixed discounting conflict repartition rule}
\label{General_rule}

From both new rules, the mixed rule (\ref{mixed}) and the discounting $\PCR$ (\ref{DPCR}), we propose a combination of these rules, given for two basic belief assignments $m_1$ and $m_2$ and for all $X \in 2^\Theta$, $X\neq \emptyset$ by:

\begin{eqnarray}
\label{MDPCR2}
\begin{array}{l}
m_\MDPCR(X)=\displaystyle  \sum_{

\begin{array}{l}
\scriptstyle Y_1\cup Y_2 =X, \\
\scriptstyle \Y_1\cap Y_2 \neq \emptyset 
\end{array}} \delta m_1(Y_1)m_2(Y_2)\\
+\displaystyle \sum_{
\begin{array}{l}
\scriptstyle Y_1\cap Y_2 =X, \\
\scriptstyle Y_1\cap Y_2 \neq \emptyset 
\end{array}} \!\!\!\!\!\!(1-\delta)  m_1(Y_1)m_2(Y_2)\\
+
\displaystyle
\sum_{
\begin{array}{l}
\scriptstyle Y\in 2^\Theta, \\
\scriptstyle X\cap Y = \emptyset 
\end{array}}\alpha \left(\frac{m_1(X)^2 m_2(Y)}{m_1(X) \!+\!
  m_2(Y)}+\frac{m_2(X)^2 m_1(Y)}{m_2(X) \!+\! m_1(Y)}\!\right)\!\!,\\

+
\displaystyle
\sum_{
\begin{array}{c}
\scriptstyle Y_1\cup Y_2 =X\\
\scriptstyle Y_1\cap Y_2 = \emptyset 
\end{array}} (1-\alpha) m_1(Y_1)m_2(Y_2).\\
\end{array}
\end{eqnarray}

$\alpha$ can be given by the equation (\ref{alpha}) and $\delta$ by the equation (\ref{delta}) or (\ref{d}). The weights are taken to get a kind of continuity between the mixed and DPCR rules. Actually, when the intersection of the responses is almost empty but not empty we use the mixed rule and when this intersection is empty we chose the $\DPCR$ rule. In the first case all the mass is transfered on the union and in the second one it will be the same according to the partial conflict. Indeed, $\alpha=0$ if the intersection is not empty and $\delta=1$ if the intersection is empty. We can also introduce $\alpha_i$ given by the equation (\ref{g_function}), and this continuity is conserved.

This rule is given in a general case for $M$ experts, for all $X \in 2^\Theta$, $X\neq \emptyset$ by:
\begin{eqnarray}
\label{eq_mixed_to_dpcr}
\begin{array}{l}

m_\MDPCR(X)=\displaystyle  \sum_{
\begin{array}{l}
\scriptstyle Y_1 \cup ... \cup Y_M=X, \\
\scriptstyle Y_1 \cap ... \cap Y_M \neq \emptyset 
\end{array}} \delta \prod_{j=1}^M m_j(Y_j)\\

+\displaystyle \sum_{
\begin{array}{l}
\scriptstyle Y_1 \cap ... \cap Y_M =X, \\
\scriptstyle Y_1 \cap ... \cap Y_M \neq \emptyset 
\end{array}} \!\!\!\!\!\!(1-\delta) \prod_{j=1}^M m_j(Y_j)\\

+
\displaystyle
\sum_{i=1}^M
  m_i(X)^2 \\
  \!\!\!\!\!\!\!\! \displaystyle \sum_{\begin{array}{c}
      \scriptstyle {\displaystyle \mathop{\cap}_{k=1}^{M\!-\!1}} Y_{\sigma_i(k)} \cap X = \emptyset \\
      \scriptstyle (Y_{\sigma_i(1)},...,Y_{\sigma_i(M\!-\!1)})\in (2^\Theta)^{M\!-\!1}
  \end{array}}
  \!\!\!\!\!\!\!\!\!\!\!\!
  \alpha \left(\!\!\frac{\displaystyle \prod_{j=1}^{M\!-\!1} m_{\sigma_i(j)}(Y_{\sigma_i(j)})}
       {\displaystyle m_i(X) \!+\! \sum_{j=1}^{M\!-\!1} m_{\sigma_i(j)}(Y_{\sigma_i(j)})}\!\!\right)\\

+
\displaystyle
\sum_{
\begin{array}{c}
\scriptstyle Y_1 \cup ... \cup Y_M = X\\
\scriptstyle Y_1 \cap ... \cap Y_M = \emptyset \\
\end{array}} (1-\alpha) \prod_{j=1}^M m_j(Y_j),
\end{array}
\end{eqnarray}
where $Y_j \in 2^\Theta$ is the response of the expert $j$, $m_j(Y_j)$ the associated belief function and $\sigma_i$ is given by (\ref{sigma}). This formula could seem difficult to understand, but we can implement it easily as shown in appendix \ref{s_algo}. 

If we take again the previous example with $\delta$ given by the equation (\ref{delta}), there is no difference with the $\DPCR$. If $\delta$ is calculated by the equation (\ref{d}), the only difference pertains to the mass 0.09 coming from the responses of the three experts: $\Theta$, $\Theta$ and $A \cup C$. This mass is transfered on $A\cup C$ (0.06) and on $\Theta$ (0.03).

\section{Discussion: toward a more general rule}
\label{illustration}

The rules presented in the previous section, propose a repartition of the masses giving a partial conflict only (when at most two experts are in discord) and do not take heed of the level of imprecision of the responses of the experts (the nonspecificity of the responses). The imprecision of the responses of each expert is only considered by the mixed and $\MDPCR$ rules when there is no conflict between the experts.

In the mixed rule, if the intersection of the responses of the experts is empty, the best way is not necessarily to transfer the mass on the union. For example, if three experts say $A\cup B$, $A\cup C$, $D$, two experts agree on $A$. So, it could be better to transfer the mass on $A$ and $A\cup B\cup C\cup D$.

Consider $M$ experts, we define the set of subsets of the responses of the experts that are not in conflict:
\begin{eqnarray}
\begin{array}{l}
\varepsilon_k(Y_1,...,Y_M)=\{ \{Y_{i_1}, ..., Y_{i_k}\}, i_j \in I:  \\
\quad \quad \quad \quad \quad I\subset \{1,...,M\}, |I|=k, \displaystyle \cap_{j=1}^k Y_{i_j} \neq \emptyset\},
\end{array}
\end{eqnarray}
where $Y_i$ is the response of the expert $i$. Additionally, we define $\overline{k}=\argmax_k\{\varepsilon_k\neq \emptyset\}$. In the previous example, $\varepsilon_{\overline{k}}=\varepsilon_3=\{ \{A\cup B, A \cup C, A\cup D\}\}$.

In the $M$ case experts, we define $\delta(Z)$ for all $Z\in \varepsilon_{\overline{k}}$ with $Z=\{Y_{i_1}, ..., Y_{i_k}\}$ as:
\begin{eqnarray}
\delta(Z)=\displaystyle 1- \frac{|\displaystyle \cap_{j=1}^{\overline{k}} Y_{i_j}|}{\displaystyle \min_{j \in \{1,...,\overline{k}\}}|Y_{i_j}|}.
\end{eqnarray}

An extended mixed rule for $M$ experts can be written:
\begin{eqnarray}
\begin{array}{l}
m_\emix(X)=\!\!\!\!\!\!\!\! \displaystyle \sum_{Y_1 \cup ... \cup Y_M=X} \;\;\sum_{Z\in \varepsilon_{\overline{k}}(Y_1,...,Y_M)} \!\!\!\!\! \!\!\!\!\!\delta (Z) \prod_{j=1}^M m_j(Y_j)\\
+ \displaystyle \!\!\!\!\!\!\!\!\!\! \sum_{
\begin{array}{l}
\scriptstyle \{Y_{i_1}, ..., Y_{i_{\overline{k}}}\}=Z \in \epsilon_{\overline{k}}(Y_1,...,Y_M)\\
\scriptstyle Y_{i_1}\cap  ... \cap Y_{i_{\overline{k}}}=X,\\
\end{array}} 
\!\!\!\!\!\!\!\!\!\! \frac{(1-\delta(Z))}{|\varepsilon_{\overline{k}}(Y_1,...,Y_M)|} \prod_{j=1}^M m_j(Y_j).
\end{array}
\end{eqnarray}

This rule keep the spirit of the mixed rule. Nevertheless, imagine a very high mass on $D$ compared to the masses on $A\cup B$ and $A\cup C$ in the previous example. Therefore, we would prefer transfer the mass proportionally on $D$ and on the other connected elements $A\cup B$ and $A\cup C$ in the spirit of $\DPCR$. For the mass allocated on these connected elements, we can apply the extended mixed rule $\emix$. Consequently, in the case of conflict between all the experts, we must find which experts are in conflict together, {\em e.g.} the connected responses of the experts. This partial conflict is more precise than the partial conflict provided considering all the responses of the experts. Thus, we obtain an extended $\MDPCR$.

To compute $\varepsilon_{\overline{k}}$ taking into account $M$ focal classes having at most a size $|\Theta|=n$, we have to read all the focal classes, and count how often each singleton appears in the focal classes: $\mathcal{O}(nM)$ operations. For each of these singletons, we might have to distribute a part of the local conflict over $k$ focal classes. Each $M$-uple of focal elements can request a treatment of $\mathcal{O}(n^2M^2)$ operations, as $k\leqslant M$ and $|\varepsilon_{\overline{k}}|\leqslant n$. If each belief function has $p$ focal elements, global complexity is bounded by $\mathcal{O}(n^2M^2p^M)$.

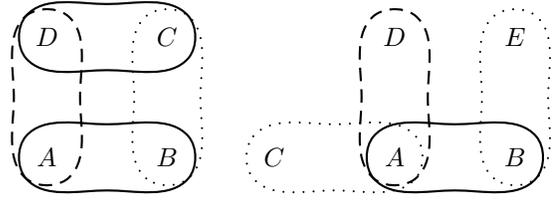
\begin{figure}
  \psset{unit=4mm}
  \begin{center}
    \begin{tabular}{cc}
      \begin{pspicture}(-3,-3)(3,3)
        \rput(-2,-2){$A$}
        \rput(2,-2){$B$}
        \rput(2,2){$C$}
        \rput(-2,2){$D$}
        \psccurve(-2.7,-2.7)(-2.7,-1.3)(0,-0.9)(2.7,-1.3)(2.7,-2.7)(0,-3.1)
        \rput{90}(0,0){\psccurve[linestyle=dotted](-2.7,-2.7)(-2.7,-1.3)(0,-0.9)(2.7,-1.3)(2.7,-2.7)(0,-3.1)}
        \rput{180}(0,0){\psccurve(-2.7,-2.7)(-2.7,-1.3)(0,-0.9)(2.7,-1.3)(2.7,-2.7)(0,-3.1)}
        \rput{270}(0,0){\psccurve[linestyle=dashed](-2.7,-2.7)(-2.7,-1.3)(0,-0.9)(2.7,-1.3)(2.7,-2.7)(0,-3.1)}
      \end{pspicture} & 
      \begin{pspicture}(-5.5,-3)(5,3)
        \rput(0,-2){$A$}
        \rput(4,-2){$B$}
        \rput(-4,-2){$C$}
        \rput(0,2){$D$}
        \rput(4,2){$E$}
        \rput{0}(2,0){\psccurve(-2.7,-2.7)(-2.7,-1.3)(0,-0.9)(2.7,-1.3)(2.7,-2.7)(0,-3.1)}
        \rput{0}(-2,0){\psccurve[linestyle=dotted](-2.7,-2.7)(-2.7,-1.3)(0,-0.9)(2.7,-1.3)(2.7,-2.7)(0,-3.1)}
        \rput{90}(-2,0){\psccurve[linestyle=dashed](-2.7,-2.7)(-2.7,-1.3)(0,-0.9)(2.7,-1.3)(2.7,-2.7)(0,-3.1)}
        \rput{90}(2,0){\psccurve[linestyle=dotted](-2.7,-2.7)(-2.7,-1.3)(0,-0.9)(2.7,-1.3)(2.7,-2.7)(0,-3.1)}
      \end{pspicture} \\
    \end{tabular}
  \end{center}
  \caption{Two conflicting focal elements sets\label{fig_epsk}}
\end{figure}

Figure \ref{fig_epsk} shows two sets of four focal elements with an empty intersection. In the left situation, each singleton is an intersection of two focal elements, and every intersection of three focal elements is $\emptyset$. So $\overline{k}$ is 2, and $\varepsilon_2$ is $\{\{A\cup B, B\cup C\}, \{B\cup C, C\cup D\}, \{C\cup D, A\cup D\}, \{A\cup B, A\cup D\}\}$.

In the right situation, $A$ appears in three focal elements, $B$ in two, and the other singletons appear only in one focal element. So $\overline{k}$ is 3, and $\varepsilon_3$ is $\{\{A\cup B, A\cup C\, A\cup D\}\}$. The singleton $E$ does not receive any part of the conflict due to its presence in the focal element $B\cup E$, but only through its presence in the disjunction of all the focal elements: $A\cup B\cup C\cup D\cup E$.

\section{Conclusions}
\label{conclusion}
In this paper, we propose some solutions to deal with the conflict and to weigh the imprecision of the responses of the experts, from the classical combination rules. Thus, we first consider a mixed rule provided by a weighted sum of the conjunctive and disjunctive rules. The weights are defined from a measure of nonspecifity calculated by the cardinality of the responses of the experts. This rule transfers the partial conflict on partial ignorance. Again, the proportional conflict distribution rule redistributes the partial conflict on the element implied in this conflict. We propose an extension of this rule by a discounting procedure, thereby, a part of the partial conflict is also redistributed on the partial ignorance. So as to quantify this part, we introduce a measure of conflict between pair of experts and another measure of non-conflict between pair of experts. In order to take heed of the nonspecifity and to redistributed the partial conflict, we propose a fused rule of these two new rules. This new rule is made in such way that we retain a kind of continuity of the mass on the partial ignorance, between both cases with and without partial conflict. Finally, we propose to discuss a more general rule that can deal with the nonspecifity of each response of the expert also in the case with partial conflict between some partition of the experts.

The comments of these new rules show that the classical combination rules in the belief functions theory cannot take precisely into account the nonspecifity of the experts and the partial conflict of the experts. We can introduce more and more artificial -or not- measures of imperfections (conflict, nonspecificity, and so on) in the conjunctive and disjunctive combination rules. Another point to treat in a futur work is how these rules perform in pratical applications.

\section{Appendix -- $\MDPCR$ algorithm}

\label{s_algo}

Formula (\ref{eq_mixed_to_dpcr}), like most of the formula of this article, seems simpler when expressed through an algorithm
instead of a direct expression of $m(X)$. We list all the $M$-uples of focal elements of the $M$ belief functions.

An input belief function $e$ is an association of a list of focal elements and their masses. We write $\size(e)$ the number of its focal elements. The focal classes are $e[1]$, $e[2]$, \ldots, $e[\size(e)]$. The mass associated to a class $c$ is $e(c)$, written with parenthesis. The cardinality of a focal element $e[i]$ is also written $\size(e[i])$.

The principle of the algorithm \ref{alg_MDPCR} is to use the variable $ind$ to build all the $n$-uples of focal elements of the $n$ input belief functions. Then, if the intersection of these is not $\emptyset$ or equivalent to $\emptyset$, the corresponding conjunctive mass (the product of all the masses of the focal elements in the $n$-uple) is put on the intersection; otherwise, it is distributed over the input focal elements and their disjunction.

\begin{algorithm}
  \caption{Fusion by the MDPCR combination rule\label{alg_MDPCR}}
  \KwData{$M$ experts $ex$: $ex[1] \ldots ex[M]$}
  \KwResult{Fusion of $ex$ by MDPCR rule : $ep$}
  \For{$i$ = 1 to $M$}{
    \ForEach{$c$ in $ex[i]$}{
      Append $c$ to $cl[i]$\;
    }
  }
  % $ind$ = [0 \ldots 0] ~~~~ // \size($ind$) = $M$.\;
  \ForEach{$ind$ in [1, $\size$($cl[1]$)] $\times$ [1, $\size$($cl[2]$)] $\times$ \ldots $\times$ [1, $\size$($cl[M]$)]}{
    $\delta$ = 1 - $\size(s)$ / $\min_{1\leqslant i\leqslant M}(\size(cl[i][ind[i]]))$\;
    $s$ $\gets$ $\Theta$; $\lprod$ $\gets$ 1; $\lsum$ $\gets$ 0; $lu$ $\gets$ $\emptyset$\;
    \For{$i$ = 1 to $M$}{
      $s$ $\gets$ $s\cap cl[i][ind[i]]$\;
      $\lprod$ $\gets$ $\lprod \times ex[i](cl[i][ind[i]])$\;
      $\lsum$ $\gets$ $\lsum + ex[i](cl[i][ind[i]])$\;        
      $lu$ $\gets$ $lu\cup cl[i]$\;
    }
    \If{$s$ = $\emptyset$}{
      $nc$ $\gets$ 0\;
      \For{$i$ = 1 to $M$}{
	\For{$j$ = 1 to $M$, $j\neq i$}{
	  \lIf{$cl[i]\cap cl[j]=\emptyset$}{$nc\gets nc+1$}\;
	}
      }
      $\alpha$ $\gets$ 1 - $nc / (M(M-1))$\;
      \For{$i$ = 1 to $M$}{
        $ep(ex[i][ind[i]]) \gets \alpha.ep(ex[i][ind[i]]) + ex[i](cl[i][ind[i]])*\lprod/\lsum$\;
      }
      $ep(lu) \gets ep(lu) + (1-\alpha)*\lprod$\;
    }
    \Else{
      $ep(s) \gets ep(s)+(1-\delta)*\lprod$\;
      $ep(lu) \gets ep(lu) + \delta*\lprod$\;
    }
  }
\end{algorithm}

% An example of a double column floating figure using two subfigures.
%(The subfigure.sty package must be loaded for this to work.)
% The subfigure \label commands are set within each subfigure command, the
% \label for the overall fgure must come after \caption.
% \hfil must be used as a separator to get equal spacing
%
%\begin{figure*}
%\centerline{\subfigure[Case I]{\includegraphics[width=2.5in]{subfigcase1}
% where an .eps filename suffix will be assumed under latex, 
% and a .pdf suffix will be assumed for pdflatex
%\hfil
%\subfigure[Case II]{\includegraphics[width=2.5in]{subfigcase2}
% where an .eps filename suffix will be assumed under latex, 
% and a .pdf suffix will be assumed for pdflatex
%\label{fig_second_case}}}
%\caption{Simulation results}
%\label{fig_sim}
%\end{figure*}

% An example of a floating table. Note that, for IEEE style tables, the 
% \caption command should come BEFORE the table. Table text will default to
% \footnotesize as IEEE normally uses this smaller font for tables.
% The \label must come after \caption as always.
%
%\begin{table}
%% increase table row spacing, adjust to taste
%\renewcommand{\arraystretch}{1.3}
%\caption{An Example of a Table}
%\label{table_example}
%\begin{center}
%% Some packages, such as MDW tools, offer better commands for making tables
%% than the plain LaTeX2e tabular which is used here.
%\begin{tabular}{|c||c|}
%\hline
%One & Two\\
%\hline
%Three & Four\\
%\hline
%\end{tabular}
%\end{center}
%\end{table}

% optional entry into table of contents (if used)
\section*{Acknowledgment}
This work is supported by the European Union (FEDER), the French state (FNADT), the Brittany region and the French Department of Finistere.

\bibliographystyle{IEEEtran}

\end{document}